\documentclass[sigconf]{acmart}
\usepackage{booktabs} 
\usepackage{tabularx} 
\usepackage[english]{babel}
\usepackage{mathrsfs}
\usepackage{multirow}
\usepackage{multicol}
\usepackage{array}
\usepackage{arydshln}
\usepackage{balance}
\usepackage{amsmath}

\usepackage{amssymb}
\usepackage{booktabs}
\usepackage[inline]{enumitem}
\usepackage{siunitx}
\usepackage{stackengine}
\usepackage{extarrows}
\usepackage{graphicx}
\usepackage{caption}
\usepackage{amsfonts,amssymb}
\usepackage{subcaption}
\usepackage{footnote}

\makesavenoteenv{tabular}
\makesavenoteenv{table}
\usepackage{xspace} 

\newcolumntype{L}[1]{>{\raggedright\arraybackslash}p{#1}}
\newcolumntype{C}[1]{>{\centering\arraybackslash}p{#1}}
\newcolumntype{R}[1]{>{\raggedleft\arraybackslash}p{#1}}

\copyrightyear{2021}
\acmYear{2021}
\setcopyright{acmlicensed}\acmConference[SIGIR '21]{Proceedings of the 44th International ACM SIGIR Conference on Research and Development in Information Retrieval}{July 11--15, 2021}{Virtual Event, Canada}
\acmBooktitle{Proceedings of the 44th International ACM SIGIR Conference on Research and Development in Information Retrieval (SIGIR '21), July 11--15, 2021, Virtual Event, Canada}
\acmPrice{15.00}
\acmDOI{10.1145/3404835.3462834}
\acmISBN{978-1-4503-8037-9/21/07}
\author{Yunxiang Zhao$^{1}$, Jianzhong Qi$^{1}$, Qingwei Liu$^{1}$, Rui Zhang$^{2,*}$}
 \affiliation{
   \institution{The University of Melbourne$^{1}$, www.ruizhang.info$^{2}$}
}
\email{{yunxiangz,qingweil}@student.unimelb.edu.au, jianzhong.qi@unimelb.edu.au, rayteam@yeah.net}
\thanks{*Corresponding author.}
\renewcommand{\shortauthors}{Yunxiang Zhao, Jianzhong Qi, Qingwei Liu, and Rui Zhang.}

\begin{document}
	
	\title{WGCN: Graph Convolutional Networks with Weighted Structural Features}
	
	\begin{abstract}
		Graph structural information such as topologies or connectivities provides valuable guidance for graph convolutional networks (GCNs) to learn nodes' representations.
		Existing GCN models that capture nodes' structural information weight in- and out-neighbors equally or differentiate in- and out-neighbors globally without considering nodes' local topologies.
		We observe that in- and out-neighbors contribute differently for nodes with different local topologies. To explore the directional structural information for different nodes, we propose a GCN model with weighted structural features, named WGCN.
		WGCN first captures nodes' structural fingerprints via a direction and degree aware Random Walk with Restart algorithm, where the walk is guided by both edge direction and nodes' in- and out-degrees. Then, the interactions between nodes' structural fingerprints are used as the weighted \textit{node structural features}.
		To further capture nodes' high-order dependencies and graph geometry,
		WGCN embeds graphs into a latent space to obtain nodes' latent neighbors and geometrical relationships.
		Based on nodes' geometrical relationships in the latent space, WGCN
		differentiates latent, in-, and out-neighbors with an attention-based geometrical aggregation.
		Experiments on transductive node classification tasks show that WGCN outperforms the baseline models consistently by up to 17.07\% in terms of accuracy on five benchmark datasets.
	\end{abstract}

	
	\maketitle

	\section{Introduction}
	\label{sec:introduction}
	Graphs are essential representations of many real-world data such as social networks, transportation networks, and e-commerce user-item graphs~\cite{lopes2016efficient,zhuang2017understanding,ren2017location,widmann2017graph,zhao2019cbhe,su2021detecting}. Graph convolutional networks (GCNs), graph attention networks (GATs), and their variants have shown promising results in applications on graph datasets~\cite{bruna2013spectral,defferrard2016convolutional,devlin2019bert,hamilton2017inductive,niepert2016learning,shuman2013emerging,lu2020vgcn,ma2020streaming,zhao2020hexcnn,xu2019incorporating}. However, most GCN and GAT based models focus on how to aggregate neighboring nodes' content features (e.g., keywords of a document in a document network) and take graph structures (edges/topology) for checking neighboring relationships only. For example, GATs learn the similarities between two nodes' content features as the attention coefficients while overlooking the relationship between nodes' local topologies (e.g., similarities).
	
	Recent studies show that explicitly adding node structural information before the aggregation procedure of GCNs substantially improves models' performance~\cite{kondor2018covariant,xu2018powerful}. However, existing approaches treat directed edges as bi-directional, i.e., neighbors connecting to a node have the same weight irrespective of the direction when capturing node structural information~\cite{pei2020geom,zhang2020adaptive,zhuang2018dual}. Such approaches ignore directional structures such as irreversible time-series relationships, which may mislead the node structural information learning~\cite{wang2020nodeaug,tong2020digraph}.
	A few studies~\cite{ma2019spectral,tong2020directed,li2020scalable,tong2020digraph}
	differentiate in- and out-neighbors when capturing node's structural information, but they give the same weight to the in-neighbors of all nodes and another the same weight to the out-neighbors of all nodes. 
	
	We believe that \emph{the weight difference between in- and out-neighbors when capturing nodes' structural information is highly related to nodes' local topologies}. 
	\begin{figure}[t]
		\setlength{\belowcaptionskip}{-0.2cm}   
		\centering
		\includegraphics[width=1\linewidth]{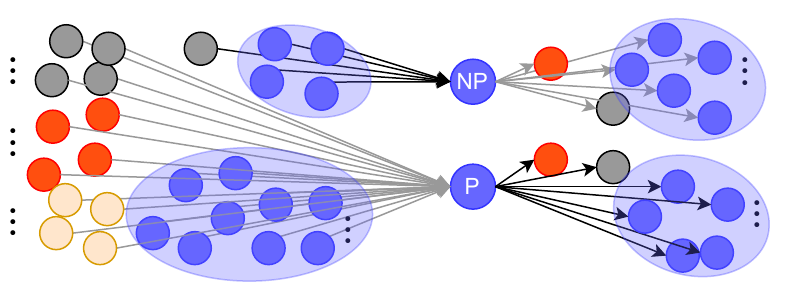}
		\caption{Citation networks for popular (P) and non-popular (NP) papers. Ellipses cover papers with the same class as the target paper. Different node colors denote different classes, such as DM, CV, and NLP (best view in color).}
		\label{fig:motivation}
	\end{figure}
	Without loss of generality, we illustrate our insight via node classification tasks for directed graphs, as shown in Figure~\ref{fig:motivation}. 
	Consider the difference between \emph{popular} and \emph{non-popular} nodes.
	A node is \emph{popular} if it has much more in-neighbors than out-neighbors, e.g., a paper with many citations but cites few other papers or a celebrity social network user with many followers but follows few other users. A \emph{non-popular} node, on the other hand, has more out-neighbors than in-neighbors, e.g., a not-so-popular paper or an average social network user. 
	In node classification, neighbors similar to the target node play an essential role, e.g., papers within the same class in citation networks and people with the same profession in social networks.
	In general, the more neighbors a node has in one direction (in or out), the higher proportion of dissimilar neighbors the node has in that direction.
	As shown in Figure~\ref{fig:motivation}, most citations of a not-so-popular paper (denoted as NP) are from the same class because, typically, only people studying in its area cite the paper. On the other hand, a popular paper (denoted as P) obtains many citations, which may come from other classes instead of the class to which the paper belongs. In summary, compared to the in-neighbors, the out-neighbors of a popular paper have a higher possibility of being similar neighbors.
	In this case, out-neighbors should carry higher weights when capturing node structural information.
	For non-popular papers, on the other hand, in-neighbors have a higher possibility of being similar neighbors, compared to the out-neighbors. In this case, in-neighbors should carry higher weights when capturing node structural information. 
	To confirm this intuition, we analyze five real datasets and compute the possibility of a node reaching a neighbor of the same class via in- or out-edges. We find that up to 21\% more nodes reach a neighbor of the same class via the direction with a smaller degree than the direction with a larger degree. This observation conforms to our intuition, which will be detailed in Section~\ref{sec:motivation}.  
	
	A straightforward strategy to differentiate the directional structural information for different nodes is to learn a parameter that weights one-hop in- and out-neighbors for each node according to its in- and out-degrees~\cite{khosla2019node}. 
	However, such a strategy cannot capture the nodes' density or differentiate nodes' geometrical relationships, and hence will lose structural information such as communities~\cite{zhang2020adaptive}. Besides, unlike continuous space such as images or videos, GCNs cannot differentiate non-isomorphic graphs due to their permutation-invariant aggregation and cannot capture long-range dependencies~\cite{pei2020geom,liu2020non}.
	Such information is critical in node representation learning. For example,
	capturing communities can help learn node structural information because nodes within the same community have closer relationships than those not within the same community~\cite{zhang2020adaptive}; capturing latent neighbors and hierarchical relationships~\cite{narayanan2016subgraph2vec,nickel2018learning} in the latent space can further improve the performance of node representation learning~\cite{hoff2002latent,muscoloni2017machine}.
	
	To address the above limitations, we propose a novel GCN variant named \emph{WGCN}. It models nodes' directional structural information, where the weights of in- and out-neighbors are determined by nodes' local topologies. WGCN contains two components. (i) the first component is a direction and degree aware Random Walk with Restart algorithm (DDRWR), where the walk is guided by both edge direction and node in- and out-degrees. We take the proximity of reaching $k$-hop neighbors from a given node as the structural fingerprint of the node. The procedure of computing the proximity quantitatively weights in- and out-neighbors for different nodes and captures communities simultaneously. Based on the structural fingerprint of each node, we define the node level interactions between node structural fingerprints to obtain an $\mathcal{N}$-dimensional vector for each node as its structural features ($\mathcal{N}$ is the number of nodes in the graph).
	(ii) the second component embeds node topology into a latent space, and nodes far away from each other in the graph may become neighbors in the latent space. Moreover, the positions of different nodes in the latent space capture their latent geometrical relationships.
	Based on nodes' geometrical relationships in the latent space, in message passing,
	WGCN performs an attention-based geometrical aggregation with learnable parameters for latent, in-, and out-neighbors, respectively.
	We summarize our contributions as follows:
	
	\begin{itemize}[leftmargin=*]
		\item We propose a novel GCN model named WGCN to capture nodes' structural information, which differentiates the weights of node in-neighbors from those of node out-neighbors according to nodes' local topologies.	
		\item We propose to embed node topology into a latent space to obtain nodes' high-order dependencies and latent geometrical relationships. We then aggregate latent, in-, and out-neighbors separately with a geometrical attention-based message passing for node representation learning. 
		\item We evaluate the proposed WGCN model with transductive node classification tasks on five public benchmark datasets, and WGCN achieves state-of-the-art performance consistently. We have made the implementation of WGCN public available\footnote{The source code is public available at: \href{https://github.com/ruizhang-ai/WGCN_Graph-Convolutional-Networks-with-Weighted-Structural-Features}{https://github.com/ruizhang-ai/WGCN\_Graph-Convolutional-Networks-with-Weighted-Structural-Features}.}.
	\end{itemize}
	
	\section{Related work}
	\label{sec:related}
	GCNs apply a message-passing strategy where each node aggregates messages/features from its neighboring nodes and updates its feature vector until an equilibrium state is reached~\cite{kipf2017semi}.
	Instead of aggregating all neighboring nodes equally, Velickovic et al.~\cite{velivckovic2018graph} propose a graph attention model (GAT) with a trainable attention function, which learns implicit information to distinguish different neighbors during the aggregation.
	Based on GAT, various approaches have been proposed for aggregating neighbors' information based on neighbors' content features and structural information~\cite{lee2019attention,wang2019heterogeneous}. 
	For example, Wang et al.~\cite{wang2019heterogeneous} propose to differentiate the weights of different neighbors for heterogeneous graphs and then aggregate content features from meta-path-based neighbors hierarchically. 
	Recently, researchers found that omitting the edge direction information will lead to information loss for node representation learning~\cite{tong2020directed}, and explicitly adding rich structural features such as the topology or "shapes" of local edge connections can further boost the performance of GCNs~\cite{zhang2020adaptive}. Therefore, various GCN models have been proposed for directed graphs, and some also explicitly capture directional structural features. They are divided into spectral and spatial approaches. 
	
	To capture the edge direction information in the spectral domain.
	Ma et al.~\cite{ma2019spectral} propose a directed Laplacian matrix for directed graphs. However, they only consider the nodes' out-neighbors (out-degree matrix) and overlook the information from in-neighbors, which may cause information loss. 
	To cope with the high computation cost of spectral-GCNs for directed graphs, Li et al.~\cite{li2020scalable} propose a scalable graph convolutional neural network with fast localized convolution operators derived from directed graph Laplacian. 
	To capture both edge direction information and nodes' local structural information, Tong et al.~\cite{tong2020directed,tong2020digraph} propose to learn first- and second-order proximities, which are combined via a signal fusion function. However, they apply the same weight for in-neighbors of all nodes and another the same weight for out-neighbors of all nodes when computing the first- and second-order proximity.
	
	Spatial GCN models mainly focus on undirected graphs when learning nodes' representations and can be applied to directed graphs by following the edge directions during the message passing.
	For example, 
	Hamilton et al.~\cite{hamilton2017inductive} propose a general inductive framework that can efficiently generate node embeddings for previously unseen data.
	Zhu et al.~\cite{zhu2019neighborhood} propose a neighborhood-aware GCN model, which considers both neighbor-level and relation-level information.
	Mostafa et al.~\cite{mostafa2020permutohedral} propose a global attention mechanism where a node can selectively attend to and aggregate content features from any other node in the graph. 
	To capture both edge direction information and nodes' local structural information,
	Zhang et al.~\cite{zhang2020adaptive} propose to use Random Walk with Restart to obtain node structural information according to node's local topology, and then train a GAT model that considers both the node content features and node structural features. 
	Spatial GCN models can be applied to directed graphs by aggregating in- or out-neighbors only during the message passing. However, considering in- or out-neighbors only may lead to information loss because neighbors in the other direction are overlooked~\cite{tong2020directed}.

\section{Methods}
\label{sec:method}
We start by analyzing real data to show that in- and out-neighbors have different probabilities of being similar neighbors for nodes with different local topologies (Section~\ref{sec:motivation}). We then present an overview of our WGCN model that captures such data characteristics in embedding learning (Section~\ref{sec:overview}). After that, we discuss how to adapt a classic algorithm -- the Random Walk -- to capture a neighbor's weight in representing a node's directional local structure information (Section~\ref{sec:theory}), detail how to compute a node's structural features based on such an algorithm (Section~\ref{sec:ddrandom}), present an attention-based geometrical message passing to support our embedding leaning (Section~\ref{sec:aggregation}), and show the time complexity of our model (Section~\ref{sec:com}). 

\subsection{Motivation and Insight}
\label{sec:motivation}
\begin{figure}[t]
	\setlength{\belowcaptionskip}{-0.2cm}   
	\centering
	\begin{subfigure}[a]{0.625\linewidth}
		\includegraphics[width=\linewidth]{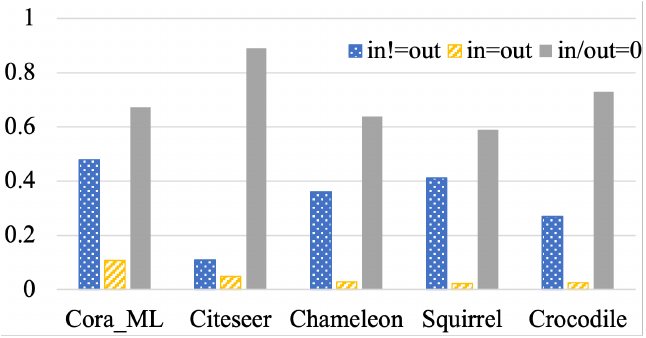} 
		\caption{}
	\end{subfigure}\,
	\begin{subfigure}[a]{0.36\linewidth}
		\includegraphics[width=\linewidth]{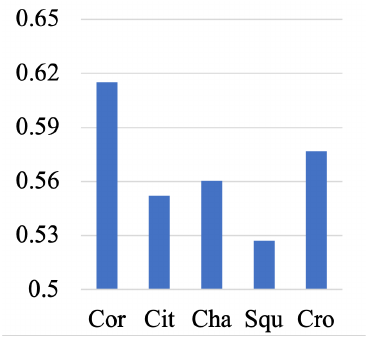}
		\caption{}
	\end{subfigure}
	\caption{(a) The percentage of nodes with different and non-zero in- and out-degrees (blue bars), the same in- and out-degrees (yellow bars), and only in- or out-neighbors (gray bars). (b) Among nodes with different and non-zero in- and out-degrees, the percentage of nodes where neighbors at the direction with a smaller degree have a higher possibility of sharing the same class with the nodes.}
	\label{fig:dataset}
\end{figure}

Our model is based on the insight that neighbors at the direction with a smaller degree have a higher possibility of being similar neighbors (e.g., being in the same class). We analyze five datasets (details in Section~\ref{sec:dataset&exp}) and report the percentage of nodes that conform to our insight.
In Figure~\ref{fig:dataset}a, the nodes in each dataset are divided into three categories: (i) nodes with in- or out-degree being zero (gray bars), i.e., nodes with only in- or out-neighbors.
(ii) nodes with the same in- and out-degrees (yellow bars).
(iii) nodes with different and non-zero in- and out-degrees (blue bars). Our model benefits the representation learning of the nodes of Category (iii), while it does not has a negative impact on nodes of the other categories. We see that the five datasets have between 52.7\% and 61.5\% of nodes of Category (iii). Improving the representations learned for such nodes is expected to bring substantial performance gains for downstream applications (e.g., node classifications, detailed in Section~\ref{sec:res}).  
Within the nodes in Category (iii), we further show the percentage of nodes that conform to our insight in Figure~\ref{fig:dataset}b. The y-axis denotes the percentage of nodes where neighbors at the direction with a smaller degree have a higher possibility of sharing the same class with these nodes. We observe that over 50\% of the nodes in Category (iii) of each dataset tested conform to our insight, confirming the necessity of weighing the in- and out-neighbors differently for the nodes. 
{For the rest nodes in Category (iii), our model has a negative impact on their representation learning, but the overall performance of all nodes is increased since they are the minority. The theoretical analysis in Section~\ref{sec:theory} and the experimental results in Section~\ref{sec:res} conform to this claim.}
Next, we present our WGCN model to learn such different weights.

\subsection{Model Overview}
\label{sec:overview}
\begin{figure*}[t]
	\setlength{\belowcaptionskip}{-0.2cm}   
	\centering
	\includegraphics[width=0.85\linewidth]{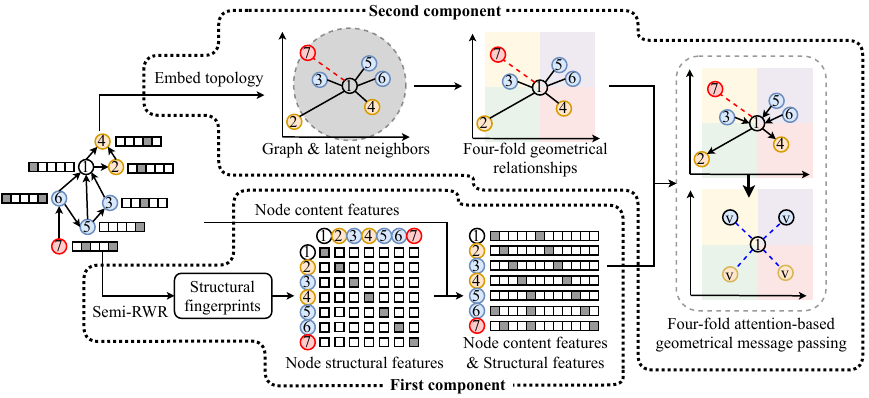}
	\caption{Overall structure of WGCN. We take 1-hop in-neighbors (blue) and out-neighbors (yellow) of node \raisebox{.5pt}{\textcircled{\raisebox{-.9pt} {1}}} in a four-fold attention-based geometrical message passing as an example. Nodes with high-order dependencies to node \raisebox{.5pt}{\textcircled{\raisebox{-.9pt} {1}}} in the latent space are red colored such as node \raisebox{.5pt}{\textcircled{\raisebox{-.9pt} {7}}} (best view in color).}
	\label{fig:struc}
\end{figure*}

The overall structure of WGCN is shown in Figure~\ref{fig:struc}, which contains two components.
Given a directed graph and node content features (e.g., keywords of a document in a document network, denoted as the vector next to each node), (i) we obtain nodes' structural fingerprints via a direction and degree aware Random Walk with Restart algorithm (DDRWR), where the walk is guided by edge direction and node in- and out-degrees. The interactions between nodes' structural fingerprints are taken as the nodes' structural features. We then concatenate structural features with node content features as the input to be passed to the message passing phase. 
(ii) meanwhile, we embed nodes into a latent space via existing embedding approaches that capture nodes' latent neighbors and the graph topology, i.e., the nodes' geometrical relationships. For each node, we differentiate its neighbors into different classes (folds) according to neighbors' geometrical relationships in the latent space.
During the message passing, we first aggregate (e.g., sum or mean) latent, in- and out-neighbors separately in each sub-space to obtain virtual nodes denoted as \raisebox{.5pt}{\textcircled{\raisebox{-.9pt} {v}}} in Figure~\ref{fig:struc}, and an attention mechanism is applied to enable nodes to discriminately attend to neighbors with different weights.
We then update node representation by aggregating (e.g., concatenate) those virtual nodes to obtain the final representation of each node. The "concatenate" aggregation can retain the order of different virtual nodes and hence can differentiate neighbors with different geometrical relationships.

\subsection{Neighbor Weight Modeling}
\label{sec:theory}
In GCN models, Random Walk has been widely used to capture structural information in node representation learning~\cite{chen2020simple,klicpera2019diffusion,zhang2020adaptive}, where the possibility of reaching each neighbor during the exploring process denotes the weight of the neighbor in representing a node's structural information. In this subsection, we first show that traditional Random Walk algorithms do not serve our purpose on directed graphs. We then present our adapted Markov process and show its advantages.

Random Walk algorithms can be seen as time-homogeneous Markov processes. Let $\{X^{[n]},~n\ge0\}$ be a time-homogeneous Markov process with state/node space $V$ and transition matrix $P=(p_{ij})_{i,j\in V}$. For any nodes $i,j\in V$:
\begin{equation} 
p_{ij}:=\mathbb{P}(X^{[n+1]}=j~|~X^{[n]}=i), ~~\forall~n\ge 0
\end{equation}
where $X^{[n+1]}$ denotes the next state generated from the current state $X^{[n]}$.
Random Walk with Restart (RWR) further captures the intuitively decaying weight of neighbors of different hops, while maintaining the community information~\cite{zhang2020adaptive}.
The RWR procedure forces the walk on a graph to always restart from the same node (or group of nodes). It can quantify the structural proximity between a given node and all other nodes in the graph~\cite{tong2006fast,zhou2017scalable}, which has been widely used for graph embedding~\cite{ou2016asymmetric,perozzi2014deepwalk,chen2020simple,klicpera2019diffusion,klicpera2018predict}.

Without loss of generality, we take graphs without self-loop and mutual connections as an example for illustration. For each node $i\in V$, we denote $A_i\subseteq V$ as the "neighbors" of node $i$, $A_i^{I}$ and $A_i^{O}$ as the in-neighbors and out-neighbors of node $i$, respectively, and $A_i=A_i^{I}\cup A_i^{O}$. Let $d_i$ be the degree of node $i$ (here and below degree means in-degree plus out-degree), i.e., $d_i:=d_i^I+d_i^O$, where $d_i^I:=|A_i^{I}|$ and $d_i^O:=|A_i^{O}|$ are the in-degree and out-degree of node $i$, respectively. 
In the traditional RWR, the possibility of any node $i\in V$ reaching next node $j$ during the exploring procedure is:
\begin{equation} 
p_{ij}:=
\begin{cases}
c,~~~&\mbox{if}~~j=o\\
\frac{1-c}{d_i},~~~&\mbox{if}~~j\in A_i\\
0,~~~&\mbox{else}
\end{cases}
\end{equation}
where node $o$ is the starting node, i.e., $\mathbb{P}(X^{[0]}=o)=1$, and $c\in(0,1)$ is a fixed constant, which indicates the possibility of restart. 

Existing studies on RWR do not consider the asymmetric roles of vertices and are not robust for directed graphs~\cite{khosla2019node}. 
Based on our insight in Section~\ref{sec:motivation}, for a node $i$, if there is a difference between its in-degree and out-degree, its neighbors in the direction with the smaller degree are more likely to be similar to node $i$. More precisely, given node $i$, if $|d_i^I-d_i^O|$ exceeds a threshold $thr\in\mathbb{N}$, for any node $j\in A_i^{I}$, and node $l\in A_i^{O}$, we have:
\begin{equation} 
\left\{
\begin{aligned}
&\mathbb{P}(j\in\mathcal{I}_i)>\mathbb{P}(l\in\mathcal{I}_i),~~&\mbox{if}~~ d_i^O-d_i^I>thr\\
&\mathbb{P}(j\in\mathcal{I}_i)<\mathbb{P}(l\in\mathcal{I}_i),~~&\mbox{if}~~ d_i^I-d_i^O>thr
\label{for:2}
\end{aligned}
\right.
\end{equation}
where $\mathcal{I}_i$ denotes all neighbors similar to node $i$, which is a random subset of $V$. In practice, we set the threshold $thr$ as 0.

Based on this insight, we propose a Markov process $\{\tilde{X}^{[n]}:~n\ge0\}$ with transition matrix $\tilde{P}=(\tilde{p}_{ij})_{i,j\in V}$, where $\tilde{p}_{ij}$ for each node $i\in V$ is computed as follows:
\begin{equation} 
\tilde{p}_{ij} = 
\begin{cases}
c, ~~&\mbox{if}~~ j=o\\
w_i^I, ~~&\mbox{if}~~ j\in A_i^{I}\\
w_i^O, ~~&\mbox{if}~~ j\in A_i^{O}\\
0,~~&\mbox{else}~~
\label{for:3}
\end{cases},~~~
\begin{cases}
w_i^I>w_i^O,~~&\mbox{if}~~d_i^O-d_i^I>thr\\
w_i^I<w_i^O,~~&\mbox{if}~~d_i^I-d_i^O>thr\\
w_i^I=w_i^O,~~&\mbox{else}
\end{cases}
\end{equation}
Here, $c+d_i^Iw_i^I+d_i^Ow_i^O=1$. 
For each node $i$, the possibility of the Markov process reaching the set of similar neighbors $\mathcal{I}_i$ is $\sum_{n\ge1}{\tilde{\mathbb{P}}}(\tilde{X}^{[n]}\in\mathcal{I}_i)$.
Throughout this paper, we only consider such a set of neighbors w.r.t. the starting node $o$. Thus, for simplification, we use $\mathcal{I}$ instead of $\mathcal{I}_o$. Assume that the distribution of $\mathcal{I}$ is independent of the distribution of $\{\tilde{X}^{[n]}:~n\ge0\}$. Then the possibility of reaching set $\mathcal{I}$ in the typical RWR and the proposed Markov process in the 1-hop neighbors are:
\begin{align*}
\mathbb{P}(X^{[n+1]}\in\mathcal{I})=\sum_{i\in V}\sum_{j\in A_i}\mathbb{P}(j\in\mathcal{I})\frac{1-c}{d_i}\mathbb{P}(X^{[n]}=i)+c
\end{align*}
\begin{align*}
\tilde{\mathbb{P}}(\tilde{X}^{[n+1]}\!\!\in\!\!\mathcal{I})\!=\!&\sum_{i\in S_1}\!\left\{\sum_{j\in A_i^I}\mathbb{P}(j\!\in\!\mathcal{I})w_i^I\!+\!\sum_{j\!\in \!A_i^o}\mathbb{P}(j\in\mathcal{I})w_i^O\right\}\tilde{\mathbb{P}}(\tilde{X}^{[n]}\!=\!i)\\
\!+\!&\sum_{i\in S_2}\!\left\{\sum_{j\in A_i^I}\mathbb{P}(j\!\in\!\mathcal{I})w_i^I\!+\!\sum_{j\in A_i^o}\mathbb{P}(j\!\in\!\mathcal{I})w_i^O\right\}\tilde{\mathbb{P}}(\tilde{X}^{[n]}\!=\!i)\\
\!+\!&\sum_{i\in S_3}\!\sum_{j\in A_i}\mathbb{P}(j\in\mathcal{I})\frac{1-c}{d_i}\tilde{\mathbb{P}}(\tilde{X}^{[n]}=i)+c\\
\end{align*}
where $n\ge0$, $S_1$, $S_2$, and $S_3$ are defined as: $S_1:=\{i\in V:~d_i^I-d_i^O>thr\}$, $S_2:=\{i\in V:~d_i^O-d_i^I>thr\}$,
$S_3:=\{i\in V:~\vert d_i^O-d_i^I\vert\le thr\}$.
When $n=0$, it is clear that $\mathbb{P}(X^{[1]}\in\mathcal{I})\le \tilde{\mathbb{P}}(\tilde{X}^{[1]}\in\mathcal{I})$. More precisely, given the initial distribution $\mathbb{P}(X^{[0]}=o)=\tilde{\mathbb{P}}(\tilde{X}^{[0]}=o)=1$, node $o$ must fall in one of $S_i$, since $\{S_1,S_2,S_3\}$ is a partition of $V$. 
If $o\in S_1$, we have
$w_o^I<\frac{1-c}{d_o}<w_o^O$ and
$\mathbb{P}(j\in\mathcal{I})<\mathbb{P}(l\in\mathcal{I})$ for any $j\in A_o^I$, $l\in A_o^O$ from Equations~\ref{for:2} and~\ref{for:3}. Therefore:  
\begin{equation}
\sum_{j\in A_o^I}\mathbb{P}(j\in\mathcal{I})w_o^I+\sum_{j\in A_o^O}\mathbb{P}(j\in\mathcal{I})w_o^O>\sum_{j\in A_o}\mathbb{P}(j\in\mathcal{I}) \frac{1-c}{d_o}\label{improvementineq}
\end{equation}
Similarly, the inequality (\ref{improvementineq}) holds for $o\in S_2$.
Because our model has the same possibility of reaching $\mathcal{I}$ as typical RWR for $o\in S_3$, then:
\begin{equation} 
\mathbb{P}(X^{[1]}\in\mathcal{I})\le \tilde{\mathbb{P}}(\tilde{X}^{[1]}\in\mathcal{I})
\end{equation}
Compared with the neighbors of a dissimilar neighbor, those of a similar neighbor have a higher possibility to be similar high-order neighbors. Therefore, with the increased possibility of reaching $\mathcal{I}$ for node $o$, the possibility of reaching $\mathcal{I}$ in high-order neighbors will also become higher in our model.

\begin{figure}[t]
	\setlength{\belowcaptionskip}{-0.2cm}   
	\centering
	\begin{subfigure}[a]{0.282\linewidth}
		\includegraphics[width=\linewidth]{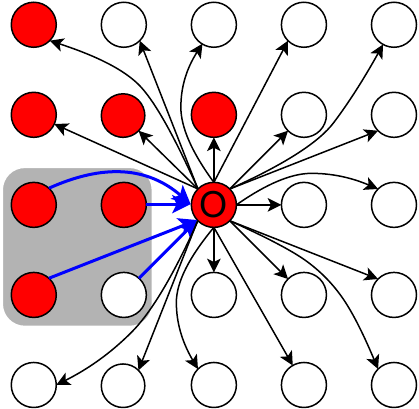}
		\caption{A simple graph.}
	\end{subfigure}
	\begin{subfigure}[a]{0.345\linewidth}
		\includegraphics[width=\linewidth]{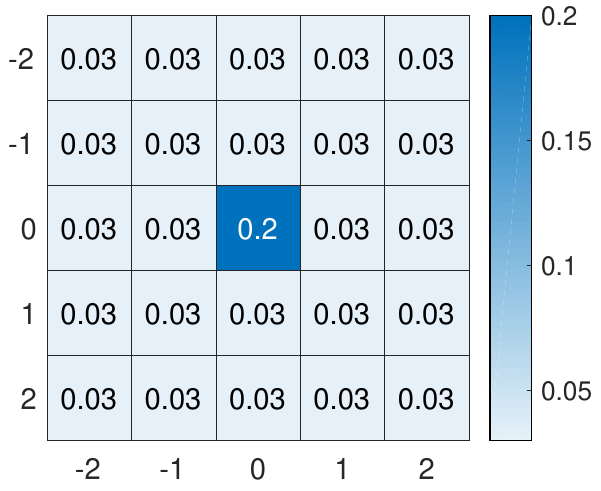}
		\caption{RWR.}
	\end{subfigure}
	\begin{subfigure}[a]{0.34\linewidth}
		\includegraphics[width=\linewidth]{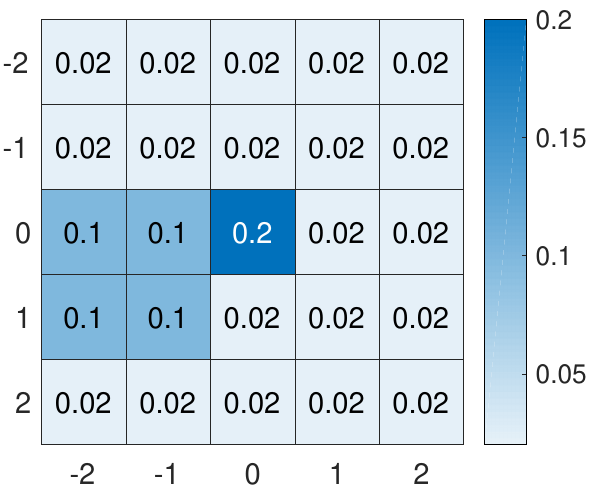}
		\caption{Ours.}
	\end{subfigure}
	\caption{Neighbor weighing via typical RWR and the proposed Markov process for node structural information learning on a simple graph.}
	\label{fig:moti}
\end{figure}

Figure~\ref{fig:moti} shows an example of using typical RWR and our Markov process to capture the node structural information. In Figure~\ref{fig:moti}a, circles denote nodes, and directed edges between nodes are the topology. Node $o$ has 4 in-neighbors and 20 out-neighbors in the graph. According to our insight, the in-neighbors should have a higher possibility to be similar neighbors. In this graph, three out of four (i.e., a probability of 75\%) in-neighbors are in the same class as node $o$. On the other hand, the out-neighbors of node $o$ have a smaller possibility to contain similar neighbors of node $o$. In Figure~\ref{fig:moti}a, there are four out of 20 (i.e., only a probability of 20\%) out-neighbors in the same class as node $o$.

If we run the typical RWR algorithm with a restart rate of 0.2 to capture the node structural information of node $o$, the weight distribution will be like Figure~\ref{fig:moti}b, where all neighbors connecting to node $o$ have the same weight. On the other hand, if we take that "in-neighbors have a higher possibility to contain similar neighbors" as pre-knowledge, and assign the in-neighbors with a higher weight, then the weight distribution will be like Figure~\ref{fig:moti}c, where the in-neighbors have a higher weight. Note that the similar neighbors of node $o$ in the out-direction have lower weights, and the dissimilar in-neighbors of node $o$ have higher weights in our method. However, the overall possibility of reaching similar nodes has been increased.

\subsection{Directional Node Structural Features}
\label{sec:ddrandom}
In GCN models, node structural features can be represented as the similarities between the local topologies of pairwise nodes. To embed the directional information into node structural features, we first present our DDRWR module for learning nodes' local topologies (fingerprints), which is based on the proposed Markov process in Section~\ref{sec:theory}. We then present the method for computing the node structural features.
\begin{figure}[t]
	\setlength{\belowcaptionskip}{-0.2cm}  
	\centering
	\begin{subfigure}[a]{0.4\linewidth}
		\includegraphics[width=\linewidth]{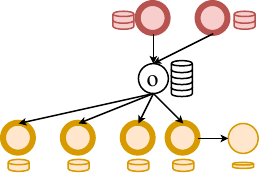}
		\caption{Neighbor weighting.}
	\end{subfigure}\,\,
	\begin{subfigure}[a]{0.56\linewidth}
		\includegraphics[width=\linewidth]{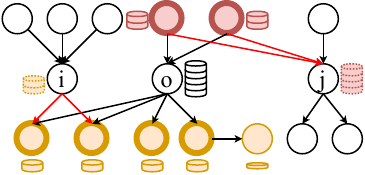}
		\caption{Structural interactions.}
	\end{subfigure}
	\caption{Neighbor weighting and node-level structural interactions in WGCN. The in-neighbors of node $o$ have higher weights due to the smaller in-degree compared to the out-degree. Node $j$ has a higher interaction weight with node $o$ because their interacted nodes have higher weights.}
	\label{fig:recep}
\end{figure}

When computing the structural fingerprint for each node, it is desirable to assign weights to neighbors adaptively based on nodes' local structures (e.g., communities). 
Figure~\ref{fig:recep}a shows a spanning process on node $o$'s $k$-hop ($k$=2 in this example) local subgraph $G_o$, and each neighbor in $G_o$ contributes a weight, which denotes the importance of the neighbor when capturing the structural fingerprint of node $o$. Based on discussions above, in-neighbors $A_o^I$ are more likely to be similar neighbors because the in-degree $d_o^I$ of node $o$ is smaller than its out-degree $d_o^O$, and hence they should have higher weights when capturing the node structural fingerprint.
Besides, even though neighbors far away from node $o$ should have smaller weights, mapping the node distance levels [1, 2, ..., k] to non-negative, monotonically decreasing weight levels $w_o = [w_1, w_2, ..., w_k ]$ will overlook the importance of communities in evaluating node similarities.

To achieve the above objectives, we combine RWR with our proposed Markov process in Section~\ref{sec:theory}, which yields an algorithm named~\emph{DDRWR}. DDRWR adjusts the walk according to the edge direction and the in/out-degrees of different nodes.
Without abuse of notations, we define the transition matrix $\tilde{P}=(\tilde{p}_{ij})_{i,j\in V}$ as:
\begin{equation}
\label{eq:4}
\tilde{p}_{ij} = 1 + b \cdot M_{ij} (\frac{ {d}_i^I- d_i^O}{ {d}_i^I+ d_i^O})^\epsilon
\end{equation}
where $b$ controls the maximum weight difference between in- and out-neighbors, $M_{ij}$ indicates the relationship between nodes $i$ and node $j$ in the graph. $M_{ij}$ equals $-1$ if node $i$ reaches node $j$ via the in-edges of node $i$ following the BFS, $M_{ij}$ equals $1$ if node $i$ reaches node $j$ via the out-edges of node $i$ following the breadth first search (BFS), and $M_{ij}$ is $0$ if node $j$ is out of reach from node $i$. Recall that ${d}_i^I$ and $d_i^O$ are the in- and out-degrees of node $i$, respectively. $\epsilon$ is zero or a positive odd number that controls the in- and out-neighbor weight variation trend regarding the in- and out-degrees ratio. For example, $\epsilon=0$ denotes that in- and out-neighbors have the same weight, $\epsilon=1$ denotes that the weight of in- and out-neighbors has a linear relationship to the in- and out-degrees' difference, and $\epsilon=3$ denotes a cubic relationship. Accordingly, we formulate the iterations of DDRWR taking node $o$ as the origin as:
\begin{equation}
\label{eq:3}
{\textbf{w}_o}^{[n+1]} = (1-c) \cdot \tilde{P}'  {\textbf{w}_o}^{[n]} + c \cdot\textbf{e}_o
\end{equation}
where $c\in[0,1]$ determines the ratio of the restart, \textbf{e$_o$} is a vector with all components equal to $0$, except the entry corresponding to node $o$, which is $1$, and $\tilde{P}'$ is the normalized version of the transition matrix $\tilde{P}$ restricted on $G_o$. 

In general, if $d_i^I$ and $d_i^O$ are close to each other, $\tilde{p}_{ij}$ approaches to the ${p}_{ij}$ in typical RWR, which means that in- and out-neighbors obtain similar weights and closer neighbors obtain a higher weight.
If $d_i^I$ is larger than $d_i^O$, then $M_{ij}$ is negative when node $i$ reaches node $j$ via its in-edges. Therefore, the weight $\tilde{p}_{ij}$ is larger for nodes reached via out-edges and smaller for nodes reached via in-edges. On the contrary, if $d_i^I$ is smaller than $d_i^O$, weight $\tilde{p}_{ij}$ is smaller for nodes reached via out-edges and larger for nodes reached via in-edges. For mutually connected neighbors of node $i$, we set their probabilities as the sum of being both in- and out-neighbor.
The converged solution of Equation~\ref{eq:3} is:
\begin{equation}
\label{eq:5}
\textbf{w}_o = (I - (1-c) \cdot \tilde{P}')^{-1}  \textbf{e}_o
\end{equation}
where $\textbf{w}_o$ quantifies the proximity of node $o$ reaching its $k$-hop neighbors $G_o$, and we take $\textbf{w}_o$ as the structural fingerprint of node $o$, $c$ controls the decay rate of the fingerprint. $\textbf{w}_o$ is zeros except for position $o$ if $c$ is $1$, and $\textbf{w}_o$ has the same distribution as that produced by a Random Walk without restart if $c$ is $0$.

After computing the structural fingerprint $\textbf{w}_o$ of each node $o$, we only keep those values of neighbors within $G_o$ (neighbors within $k$-hops) and normalize $\textbf{w}_o$ to compute the node level structural interactions. We use weighted Jaccard similarity to evaluate the structural interactions ${S}=({s}_{ij})_{i,j\in V}$ as follows:
\begin{equation}
\label{eq:6}
s_{ij} = \frac{\sum_{g\in(G_i\cup G_j)}min({w}_{ig},{w}_{jg})}{\sum_{g\in(G_i\cup G_j)}max({w}_{ig},{w}_{jg})}
\end{equation}
where $w_{ig}$ and $w_{jg}$ are the $g$th values of the normalized $\textbf{w}_i$ and $\textbf{w}_j$, respectively.
Each $s_{ij}$ in $S$ is a value denoting the structural interaction between node $i$ and $j$. Figure~\ref{fig:recep}b illustrates Equation~\ref{eq:6}. For node $o$, out-neighbors have a higher weight because the out-degree is smaller than the in-degree. Nodes $i$ and $j$ each has two intersected neighbors with node $o$, but node $o$ has a higher structural interaction with node $j$ because of the higher weights of their intersected neighbors.

\subsection{Geometrical Message Passing}
\label{sec:aggregation}
\begin{figure}[t]
	\setlength{\belowcaptionskip}{-0.2cm}   
	\includegraphics[width=1\linewidth]{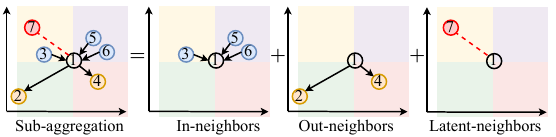}
	\caption{An example of four-fold sub-aggregation. }
	\label{fig:hmp}
\end{figure}

To explore node latent geometrical relationships, we embed each node $i$ in the graph by $f:i\rightarrow \textbf{z}_i$, where $\textbf{z}_i$ denotes the position of node $i$ in the latent space, function $f$ is an embedding function that preserves node latent geometrical relationships. We apply Isomap embedding later in the experiments, which can be replaced by other approaches that preserve node latent geometrical relationships.
In the latent space, complex topology patterns can be preserved and presented as intuitive geometry, such as subgraph, community, and hierarchy~\cite{narayanan2016subgraph2vec,ni2019community,nickel2017poincare,pei2020geom}.
We apply a distance threshold to determine the latent neighbors of each node, as the dashed circle shown in Figure~\ref{fig:struc}. The way to obtain the threshold is the same as that in Geom-GCN~\cite{pei2020geom}. To capture nodes' geometric relationships, for node $i$'s neighbors $N_i$ in the latent space ($G_i \subseteq N_i$ because $N_i$ may contain latent neighbors), WGCN computes their relative positions:
\begin{equation}
\label{eq:9}
\tau(\textbf{z}_i,\textbf{z}_j) \rightarrow r \in R, \,\,\,j \in N_i
\end{equation}
where $\tau$ is a function defined in the latent space, and $R$ is a set of relationships. For example, in 2D embedding space, a four-fold relationship set $R$ includes 13 relationships as shown in Figure~\ref{fig:hmp}: upper-left, upper-right, lower-left, lower-right for latent, in- and out-neighbors, respectively, and one last relationship is self-loop. Given the positions of two nodes in 2D embedding space, the function $\tau$ in Equation~\ref{eq:9} will return one of the 13 relationships.

After obtaining the geometrical relationships for all neighbors $N_i$ of a given node $i$, WGCN either performs attention-based or pure sub-aggregation. For attention-based sub-aggregation, WGCN first concatenates node structural features to content features as the new features of each node $i$ via $\textbf{h}_{i}=(\tilde{\textbf{x}}_i||\tilde{\textbf{s}}_i)\in\mathbb{R}^d$, where $\tilde{\textbf{x}}$ and $\tilde{\textbf{s}}$ are two vectors denoting the row-normalized (described in GCN) content features $\textbf{x}_i$ and structural features $\textbf{s}_i$ of node $i$, $||$ denotes the concatenate operator. 
For neighbors with each type of relationship $r$, WGCN learns a projection matrix $W_r\in \mathbb{R}^{d\times d}$ parameterizing a particular attention function $\textbf{A}\in\mathcal{A}$ to enable the sub-aggregation to discriminate the neighbors with relationship $r$ to node $i$. The unnormalized attention coefficient between node $i$ and its neighbors in one of the sub-aggregation spaces $r$ is:
\begin{equation}
\label{eq:7}
e_{ij}=\textbf{A}(W_r\textbf{h}_i,W_r\textbf{h}_j), \,j\in N_{i,r} 
\end{equation}
where $e_{ij}$ denotes the importance of node $j$ to node $i$, $N_{i,r}$ denotes the set of neighbors with relationship $r$ to node $i$ only, i.e., $$N_{r,i}:=\{j\in N_i:~j\overset{r}{\to} i\}.$$ 

We then apply a Softmax function to $e_{ij}$, and the attention coefficient between each pair of nodes $i$ and its neighbor $j$ is:
\begin{equation}\label{eq:2}
\alpha_{ij}=\frac{exp(LeakyRelu(e_{ij}))}{\sum_{g\in N_{i,r}} exp(LeakyRelu(e_{ig}))}
\end{equation}
where $\alpha_{ij}$ denotes the coefficient between node $i$ and node $j$. With the attention coefficients $\alpha$, we summarize the sub-aggregation of node $i$ under relationship $r$ in WGCN as:
\begin{equation} 
\label{eq:10t}
\begin{split}
\textbf{v}_{i,r} = ||_{\textbf{A}\in \mathcal{A}}\sum_{g\in N_{i,r}}\alpha_{ig} W_r\textbf{h}_{g}
\end{split}
\end{equation}
where $\textbf{h}_{g}$ denotes the features of node $g$. 
The sum operator can be replaced by other permutation-invariant operators such as the mean operator.
The attention function $\textbf{A}$ (with $\alpha_{ig}$ and $W_r$) is learned for each relationship $r$. $||_{\textbf{A}\in \mathcal{A}}$ concatenates the output from all attention heads $\mathcal{A}$ to obtain the features of virtual node $\textbf{v}_{i,r}$. The corresponding equation to obtain features of virtual node $\textbf{v}_{i,r}$ in pure sub-aggregation can be easily inferred from Equation~\ref{eq:10t}.
Based on the virtual nodes from neighbors with different relationships, WGCN aggregates the features of all virtual nodes as follows:
\begin{equation} 
\label{eq:11}
\begin{split}
\textbf{h}'_i = \sigma(\hat{W}Q([\textbf{v}_{i,r_1},...\textbf{v}_{i,r_n}]))
\end{split}
\end{equation}
where $\hat{W}$ denotes a learnable weight matrix for the overall aggregation, $\sigma$ denotes a non-linear activation function, function $Q$ is a permutation-variant function (e.g., concatenate) that summarizes the features from all virtual nodes, so as to explore the neighbors' geometry relationships. 
Equations~\ref{eq:10t} and~\ref{eq:11} work together as one message passing layer in WGCN, and multi-layer message passing can be implemented by aggregating Equations~\ref{eq:10t} and~\ref{eq:11} iteratively.

\subsection{Complexity Analysis}
\label{sec:com}
We analyze the complexity of WGCN as follows. To compute the structural fingerprint of each node, we reach $k$-hop neighbors $G_i$ via BFS, where $k$ is two for assortative datasets and one for disassortative datasets in our experiments. The complexity is $O(\mathcal{N} \cdot \mathcal{K})$, where $\mathcal{N}$ denotes the number of nodes in the graph, and $\mathcal{K}$ denotes the average number of $k$-hop neighbors. For each node, we compute the interactions with its $2k$-hop neighbors via Jaccard similarity, and the complexity is $O(\mathcal{K})$. Therefore, the overall complexity for learning node structural features is still $O(\mathcal{N} \cdot \mathcal{K})$. The computation of node structural features is a one-off pre-processing step before the training process.
During the message passing, WGCN combines node structural features with node content features and then performs an attention-based geometrical aggregation. Take WGCN without attention mechanism as an example (adding attention mechanism does not change the time complexity~\cite{velivckovic2018graph}), the time complexity is $O((\mathcal{N}$+$\mathcal{F})\cdot \mathcal{E})$, where $\mathcal{F}$ denotes the dimensionality of the original node content features, and $\mathcal{E}$ denotes the number of edges in the graph. An important future work is to develop accelerating technology for improving the scalability of WGCN.

\section{Experiments}
\label{sec:experiments}
We compare WGCN with state-of-the-art models on graph datasets for transductive node classification tasks.

\subsection{Datasets and Experimental Setup}
\label{sec:dataset&exp}
\textbf{Datasets:} We use both assortative and disassortative graph datasets. Assortative graphs~\cite{liu2020non} refer to those with high node homophily, and we use two directed citation graphs, Cora-ML~\cite{bojchevski2018deep} and Citeseer~\cite{bojchevski2018deep}. The graph nodes represent articles, while the edges represent citations between articles. Both datasets also include bag-of-words feature vectors for each article.
Disassortative graphs~\cite{liu2020non} contain more nodes that share the same class labels but are distant from each other. We use three directed Wikipedia page link graphs
Chameleon, Squirrel, and Crocodile. These are page-to-page link networks on three different topics. In these datasets, the nodes represent web pages, and the edges are directed links from one page to another. The nodes' content features correspond to informative nouns in the Wikipedia pages. The nodes are evenly divided into five classes according to their average monthly traffic (number of visits)~\cite{rozemberczki2019multi}. 
Table~\ref{tab:stat} summarizes these datasets.
\begin{table}[t]
	\centering
	\caption{Datasets statistics.}
	\vspace{-0.4cm}
	\label{tab:stat}
	\setlength{\tabcolsep}{2pt}
	\begin{tabular}{lccccccc}
		\toprule
		\multicolumn{1}{l}{\textbf{Dataset}}  &Cora-ML  &Citeseer    &  Chameleon   &Squirrel & Crocodile\\
		\midrule
		Nodes     &2,995  &4,320      &2,277   &5,201 & 11,631\\
		Edges     &8,416  &5,358      &36,101    & 217,073     &180,021 \\
		features  &2,879  &602        & 2,325  & 2,089    &13,183 \\
		Classes   &7      & 6         & 5  & 5    & 5\\
		Assortative &Yes  &Yes        &No  & No    & No \\
		\bottomrule
	\end{tabular}
	\vspace{-0.2cm}
\end{table}

\textbf{Experimental setup:}
\label{sec:setup}
We implement a two-layer WGCN.
For each layer, we first apply a four-fold geometrical aggregation to obtain virtual nodes that capture the latent geometry relationships. To capture node latent geometrical relationships, we apply the Isomap embedding~\cite{tenenbaum2000global}, where the distance patterns (lengths of shortest paths) and geometrical information are preserved in the latent space. We then concatenate the output of different virtual nodes as the output. 
During the structural fingerprint generation, we set the hops of neighbors to consider as two during the DDRWR exploring for assortative graphs and one for disassortative graphs, and $\epsilon$ as three.
Parameter $b$ and $c$ are dataset and embedding algorithm related. We use the grid search to find optimal values for them in the range of [0,1] with a step length of 0.1. In general, setting $b\in[0.2, 0.4]$ and $c\in[0.4,0.6]$ yields a satisfactory performance.
We set the parameter $b$ as 0.3 and $c$ as 0.5 for all datasets. 
We use Adam optimizer~\cite{kingma2015adam}, and ReLU as the activation function for the overall aggregation.
We run our model for 500 epochs, use a dropout ratio of 0.5, a learning rate of 0.05, and a weight decay of 5e-6. We set the number of attention head as one, and the number of hidden units as 48 for both layers.

\begin{table*}[tbh]
	\centering
	\caption{Mean Classification Accuracy (Percent) and Training Time for 100 Epochs (Second).}
	\vspace{-0.3cm}
	\setlength{\tabcolsep}{1.0mm}
	\label{tab:resl}
	\begin{tabular}{lcc|cc|cc|cc|cc}
		\toprule
		\multirow{3}{*}{\textbf{Models}} 
		&\multicolumn{4}{c|}{\textbf{Assortative}} &\multicolumn{6}{c}{\textbf{Disassortative}}\\
		&\multicolumn{2}{c|} {Cora-ML} & \multicolumn{2}{c|} {Citeseer} & \multicolumn{2}{c|} {Chameleon}  &\multicolumn{2}{c|} {Squirrel}  &\multicolumn{2}{c} {Crocodile} \\
		\cmidrule{2-11}
		& Acc & Time & Acc & Time & Acc & Time & Acc & Time & Acc & Time\\ 
		\midrule
		GCN~\cite{kipf2017semi}&76.23$\pm$1.3&4.69&84.40$\pm$0.9&4.00&32.21$\pm$2.1  &13.51&25.69$\pm$1.6 &46.27&55.87$\pm$0.8&41.15  \\
		DGCN~\cite{zhuang2018dual} &30.07$\pm$5.4&15.80&95.22$\pm$0.8&5.60&61.84$\pm$1.4  &11.26&20.27$\pm$0.7 &40.65&19.82$\pm$0.9&135.99  \\
		H-GCN~\cite{hu2019hierarchical} &72.60$\pm$4.4  &9.97& 91.59$\pm$1.2 &8.32&42.32$\pm$0.1  &8.12&23.54$\pm$0.1   &38.47&44.01$\pm$0.1&46.52 \\
		PH-GCN~\cite{mostafa2020permutohedral}&59.25$\pm$3.4&2.56& 80.47$\pm$2.4 &2.05&30.70$\pm$3.6 &3.94&21.13$\pm$0.9 &17.70&55.09$\pm$0.9&42.96 \\
		PH-GCN\_{ea}~\cite{mostafa2020permutohedral}&64.95$\pm$2.5&51.70& 83.65$\pm$2.1 &87.17&35.53$\pm$3.4 &44.42&24.11$\pm$1.1
		&58.06&57.41$\pm$1.1&115.68 \\
		DiGCN~\cite{tong2020digraph} &81.22$\pm$0.5 &13.14&85.14$\pm$0.5 &12.04&51.92$\pm$1.5 &14.71&34.32$\pm$1.1 &38.23&65.47$\pm$0.3&75.77 \\
		DiGCN\_ib~\cite{tong2020digraph} &83.55$\pm$0.2 &22.28&89.33$\pm$0.5 &21.60&50.59$\pm$1.9 &25.86&33.67$\pm$0.6 &92.57&63.91$\pm$0.3&170.59 \\
		GAT~\cite{velivckovic2018graph}&69.87$\pm$2.2&30.25&79.90$\pm$1.3&25.03&38.09$\pm$1.5  &96.35&27.89$\pm$1.4  &358.00&56.67$\pm$1.3&315.84 \\
		ADSF-RWR~\cite{zhang2020adaptive}&70.92$\pm$2.6 &30.02&80.06$\pm$1.2&25.11&40.65$\pm$1.8  &95.71&29.01$\pm$1.2   &357.53&57.37$\pm$1.3&304.64 \\
		Geom-GCN${\rm_{I}}$~\cite{pei2020geom}&80.94$\pm$1.4&28.23&92.77$\pm$0.8&22.08&60.41$\pm$3.0 &133.79&32.23$\pm$2.2  &793.29& 62.37$\pm$0.8 &441.84\\
		Geom-GCN${\rm_{S}}$~\cite{pei2020geom}&78.59$\pm$1.2&37.84& 91.50$\pm$0.8 &31.37&59.54$\pm$3.1  &138.17&34.45$\pm$1.3  &482.75&57.63$\pm$0.4&427.52 \\
		Geom-GCN${\rm_{P}}$~\cite{pei2020geom}&82.06$\pm$1.3&31.30& 94.96$\pm$0.7 &27.27 &53.46$\pm$2.0&202.74 &35.98$\pm$1.3   &428.69&  47.47$\pm$1.1&419.32\\
		\midrule
		WGCN &\textbf{87.31$\pm$2.1}&27.42&\textbf{96.45$\pm$0.8} &22.45&\textbf{67.25$\pm$2.9} &74.97&\textbf{53.05$\pm$3.8}  &425.56&  \textbf{75.57$\pm$0.6}&580.13 \\
		WGCN${\rm_{S}}$ &{86.83$\pm$2.0} &29.88& {95.75$\pm$1.0} &22.25&66.38$\pm$7.9 &76.40&50.75$\pm$2.6  &419.82&{75.49$\pm$0.5}&578.45 \\
		WGCN${\rm_{P}}$ &86.99$\pm$1.2&27.04& 95.84$\pm$1.1 &22.50&62.59$\pm$3.3 &74.94& {45.81$\pm$2.2}    &425.96& 73.12$\pm$0.9&594.12\\
		\bottomrule
	\end{tabular}
\end{table*}

\textbf{Baselines:}
We compare our model with nine state-of-the-art models divided into three groups: 1) spectral-based GNNs including GCN~\cite{kipf2017semi}, DGCN~\cite{zhuang2018dual}, and H-GCN~\cite{hu2019hierarchical}; 
2) spatial-based GNNs containing GAT~\cite{velivckovic2018graph}, ADSF~\cite{zhang2020adaptive}, Geom-GCN~\cite{pei2020geom} (Geom-GCN${\rm_{I}}$, Geom-GCN${\rm_{S}}$, and Geom-GCN${\rm_{P}}$ are its variants with different embedding methods), and PH-GCN~\cite{mostafa2020permutohedral} (PH-GCN$\_$ea replaces the original GAT attention mechanism in PH-GCN by Euclidean attention mechanism); 
3) DiGCN~\cite{tong2020digraph}, which is the SOTA GNN model for directed graphs. The source codes of other GNN models for directed graphs~\cite{li2020scalable,ma2019spectral,tong2020directed} are not publicly available.
For GCN and GAT, we follow their implementations in the Geom-GCN paper. For all other baselines, we set their parameters the same as those in the original paper.
We randomly split the datasets and perform 10 experiments for each model to obtain more reliable results.
For all baselines and our WGCN model, we randomly split nodes of each class into 60\%, 20\%, and 20\% for training, validation, and testing, respectively. This data splitting has been used in existing works such as Geom-GCN and PH-GCN. 
We run experiments with NVIDIA Tesla P100~\cite{lafayette2016spartan}.

\subsection{Results}
\label{sec:res}
The experimental results are summarized in Table~\ref{tab:resl}, where the reported values denote the mean classification accuracy in percentage. 
WGCN achieves state-of-the-art performance on both assortative and disassortative datasets.
Specifically, (i) on assortative datasets, WGCN achieves state-of-the-art performance on the Cora-ML and Citeseer datasets with 3.76\% and 1.22\% accuracy improvements, respectively;
(ii) on disassortative datasets, WGCN outperforms baselines by up to 5.41\% on the Chameleon dataset, 17.07\% on the Squirrel dataset, and 10.10\% on the Crocodile dataset. 

To analyze the importance of the embedding methods for capturing nodes' latent neighbors, we implement two more variants of WGCN with Struc2vec~\cite{ribeiro2017struc2vec} (WGCN${\rm_{S}}$) and Poincare~\cite{nickel2017poincare} (WGCN${\rm_{P}}$) embedding methods, respectively. Struc2vec embedding that capturing node structure (k-hop neighbors' degree) similarities has a competitive performance on all datasets except the Squirrel dataset, where the accuracy is 2.3\% lower than that of WGCN with Isomap embedding~\cite{tenenbaum2000global}. Its inferior performance may be due to the lack of capturing nodes' geometrical relationships, which does not suit our WGCN model. The performance of Poincare embedding achieves competitive performance on citation datasets. However, it achieves less satisfactory results on Wikipedia page link datasets with up to 7.24\% drops in the accuracy. This is because Poincare embedding assumes that nodes have latent hierarchical relationships (e.g., mammal and rodent), which may not hold on the Wikipedia page link datasets on a specific topic. Therefore, we recommend using Isomap embedding (or other embedding methods) that can capture the nodes' geometrical relationships for our WGCN model.

\subsection{Efficiency}
\label{sec:effi}
\begin{figure}[t]
	\setlength{\belowcaptionskip}{-0.2cm}    
	\centering
	\includegraphics[width=\linewidth]{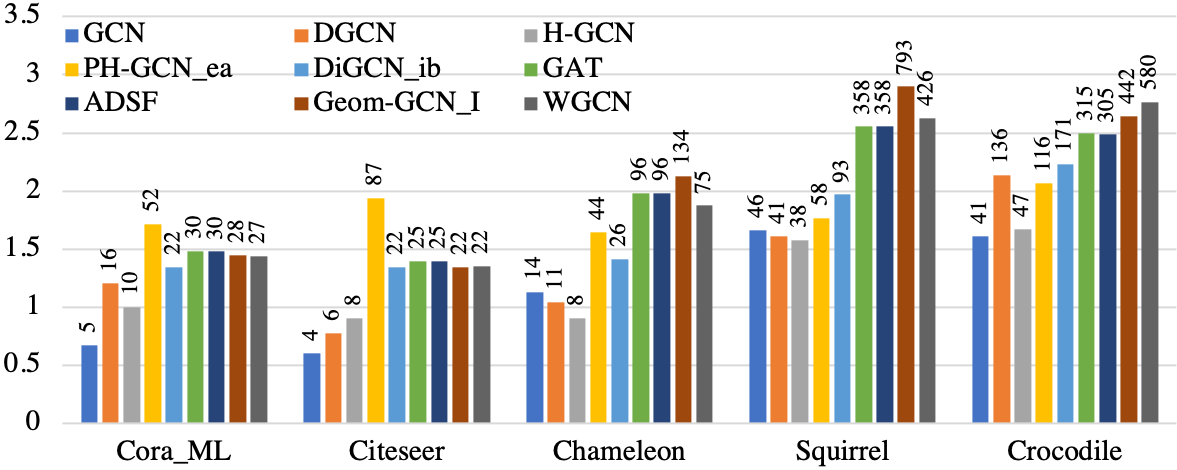}
	\caption{The time costs of different models on different datasets in 100 epochs. {The y axis is the running time in the Log scale, and the labels are the running times in seconds.}}
	\label{fig:com}
\end{figure}

To evaluate the time costs of WGCN, we run each model with 100 epochs on different datasets and summarize the running times in Figure~\ref{fig:com}. 
On assortative datasets, the running time of WGCN is on par with 5 out of 8 baselines, except for the GCN, DGCN, and H-GCN models. For disassortative graphs, WGCN still achieves comparable running times with DiGCN\_ib, GAT, ADSF, and Geom-GCN. Considering the performance gain of WGCN, we argue that its extra time costs are worthwhile.

\subsection{Ablation Study}
WGCN contains two main components. The first component DDRWR applies a direction and degree aware RWR to capture the rich directional structural information. To study its impact, we implement a variant of WGCN named WGCN/E that takes all neighbors with the same weights when capturing the node structural information. In other words, we replace our proposed DDRWR with typical RWR to capture the node structural information. We skip variants using vanilla Random Walk~\cite{perozzi2014deepwalk} as RWR has been shown to capture node structural features better~\cite{zhang2020adaptive}. The second component of WGCN captures nodes' geometrical relationships and latent neighbors for node structural information learning. 
To study its impact, we implement two variants of WGCN named WGCN/G and WGCN/L. WGCN/G does not differentiate the geometrical relationships of different neighbors. In WGCN/L, the latent neighbors are removed during the geometrical aggregation (e.g., node \raisebox{.5pt}{\textcircled{\raisebox{-.9pt} {7}}} in Figure~\ref{fig:struc}).

\begin{table}[tbh]
	\centering
	\caption{Mean Classification Accuracy (Percent).}
	\vspace{-0.4cm}
	\label{tab:abl}
	\setlength{\tabcolsep}{1.3mm}
	\begin{tabular}{lcc|ccc}
		\toprule
		\multirow{2}{*}{\textbf{Models}} 
		&\multicolumn{2}{c|}{\textbf{Assortative}} &\multicolumn{2}{c}{\textbf{Disassortative}}\\
		&Cora-ML & Citeseer& Chameleon  &Squirrel \\
		\midrule
		WGCN &{87.31$\pm$2.09} & \textbf{96.45$\pm$0.8} &{67.25$\pm$2.98} &\textbf{53.05$\pm$3.83}    \\
		WGCN/G &\textbf{87.49$\pm$0.86}&96.41$\pm$0.3&\textbf{67.43$\pm$5.42}  &52.95$\pm$4.26   \\
		WGCN/L &87.01$\pm$1.56 &96.41$\pm$0.6 &67.19$\pm$3.70  &52.63$\pm$1.28   \\
		WGCN/E &86.78$\pm$0.92 &96.22$\pm$0.8 &66.71$\pm$5.48  &51.97$\pm$1.76    \\
		\bottomrule
	\end{tabular}
	\vspace{-0.1cm}
\end{table}

The results of different variants are summarized in Table~\ref{tab:abl}. In general, WGCN and WGCN/G achieve similar accuracy with at most 0.18\% difference. Meanwhile, WGCN achieves lower variances on the disassortative datasets, and WGCN/G achieves lower variances on the assortative datasets.
Therefore, we recommend using WGCN for disassortative datasets and WGCN/G for assortative datasets. The geometrical relationships among assortative datasets are less important than that of disassortative datasets. 
For WGCN/E and WGCN/L, we see that they are consistently worse than or equal to WGCN and WGCN/G. This confirms that differentiating the in- and out-neighbors and capturing the long-range dependencies are important when capturing node structural information.

\subsection{Parameter Study}
\label{sec:ablation}
We examine the hyper-parameters $\epsilon$, the number of hops of neighbors $k$ during the DDRWR process, and the number of network layers of WGCN. We set the default values of $\epsilon$ as 3, $k$ as 2 for assortative datasets and 1 for disassortative datasets, and the number of network layers as two.

\textbf{Impact of $\epsilon$:}
Parameter $\epsilon$ determines the weight variation trend of in- and out-neighbors under a given ratio between the in- and out-degrees. A smaller value of $\epsilon$ denotes that the weight is more sensitive to the ratio (e.g., linear relationship to the ratio when $\epsilon$ is 1).
A larger value of $\epsilon$ denotes that the weight is less sensitive to the ratio but changes sharply if neighbors are mostly in- or out-neighbors (e.g., the variation trend of quintic function). We report the performance of WGCN with $k \in [1,9]$ and a step length of 2. The results in Table~\ref{tab:eps} show that WGCN performs the best when the value of $\epsilon$ is 3 for all datasets. When $\epsilon>3$, the model's performance decreases on all datasets slightly. Therefore, we take $\epsilon=3$ as a choice for yielding a good performance.
\begin{table}[tbh]
	\centering
	\vspace{-0.1cm}
	\caption{Accuracy with different $\epsilon$.}
	\vspace{-0.4cm}
	\label{tab:eps}
	\setlength{\tabcolsep}{1.3mm}
	\begin{tabular}{lcc|ccc}
		\toprule
		\multirow{2}{*}{\textbf{$\epsilon$}} 
		&\multicolumn{2}{c|}{\textbf{Assortative}} &\multicolumn{2}{c}{\textbf{Disassortative}}\\
		&Cora-ML & Citeseer& Chameleon  &Squirrel \\
		\midrule
		1	&86.49$\pm$1.63	&96.22$\pm$0.33	&{67.14$\pm$5.20}	& 51.89$\pm$2.26\\
		3	&\textbf{87.31$\pm$2.09}	&\textbf{96.45$\pm$0.80}	&\textbf{67.25$\pm$2.98} & \textbf{53.05$\pm$3.83}\\	
		5	&85.60$\pm$4.29	&96.18$\pm$0.40	&66.78$\pm$2.78 & 52.21$\pm$1.25\\	
		7	&86.41$\pm$1.13	&96.16$\pm$0.49	&66.03$\pm$4.09 & 52.93$\pm$1.43\\ 	
		9	&86.46$\pm$3.05	&96.16$\pm$0.45	&66.97$\pm$4.49 & 51.49$\pm$2.23\\	
		\bottomrule
	\end{tabular}
\end{table}

\textbf{Impact of $k$:}
The neighbors' range determines the neighbors that WGCN considers when generating node structural fingerprint. We report the performance of WGCN with $k$-hop neighbors, where $k \in [1,3]$. The results in Table~\ref{tab:khops} show that WGCN achieves the highest accuracy using 2-hop local neighbors for assortative graphs and 1-hop local neighbors only for disassortative graphs. This possibly is because, in assortative graphs, similar nodes tend to be close to each other, where 2-hop neighbors still maintain a strong relationship with the target node, while this relationship is weaker on disassortative graphs. 
Notice that the $k$ determines the hops of graph neighbors to consider during the message passing, our model will also involve neighbors that close in the latent space but beyond $k$-hop in the graph topology.
\begin{table}[t]
	\centering
	\caption{Accuracy with different $k$.}
	\vspace{-0.4cm}
	\label{tab:khops}
	\setlength{\tabcolsep}{1.3mm}
	\begin{tabular}{ccc|ccc}
		\toprule
		\multirow{2}{*}{\textbf{$k$-hops}} 
		&\multicolumn{2}{c|}{\textbf{Assortative}} &\multicolumn{2}{c}{\textbf{Disassortative}}\\
		&Cora-ML & Citeseer& Chameleon  &Squirrel \\
		\midrule
		1	&85.85$\pm$1.96 &95.90$\pm$0.78	&\textbf{67.25$\pm$2.98}	&\textbf{53.05$\pm$3.83} \\
		2	&\textbf{87.31$\pm$2.09}	&\textbf{96.45$\pm$0.80}	&66.07$\pm$3.47	& 48.99$\pm$1.56\\
		3	&86.03$\pm$0.63	&95.86$\pm$0.55	&65.15$\pm$1.89	& 47.98$\pm$5.44\\
		\bottomrule
	\end{tabular}
	\vspace{-0.1cm}
\end{table}

\textbf{Impact of the number of network layers:}
As shown in Section~\ref{sec:aggregation}, our model first aggregates the nodes with different relationships to the target node into different virtual nodes and then aggregates all the virtual nodes to obtain the final representation of the target nodes. We add multiple layers (each layer contains a local and then a global aggregation) to evaluate the impact of the number of network layers on the model performance. As summarized in Table~\ref{tab:nl}, when there are more layers, our model's performance decreases in general. 
We see that using two layers yields the best performance on both assortative and disassortative graphs. 
More specifically, on the assortative graphs, the performance of a 3-layer WGCN yields good performance, but the performance drops when the number of the network layers increases to four. On disassortative graphs, the performance of a 3-layer WGCN decreases a lot compared with a 2-layer WGCN, and the performance is even worse when the number of network layers increases to four. 
\begin{table}[tbh]
	\centering
	\vspace{-0.1cm}
	\caption{Accuracy with different network layers.}
	\vspace{-0.4cm}
	\label{tab:nl}
	\setlength{\tabcolsep}{1.3mm}
	\begin{tabular}{ccc|ccc}
		\toprule
		\multirow{2}{*}{\textbf{Layers}} 
		&\multicolumn{2}{c|}{\textbf{Assortative}} &\multicolumn{2}{c}{\textbf{Disassortative}}\\
		&Cora-ML & Citeseer& Chameleon  &Squirrel \\
		\midrule
		2 &\textbf{87.31$\pm$2.09} & \textbf{96.45$\pm$0.8} &\textbf{67.25$\pm$2.98}  &\textbf{53.05$\pm$3.83}    \\
		3 &83.79$\pm$3.56 & 95.84$\pm$0.45 &58.53$\pm$3.95  &{33.79$\pm$2.86}    \\
		4 &47.81$\pm$6.44 & 90.32$\pm$1.63 &41.62$\pm$10.05  &{20.91$\pm$4.26}    \\
		\bottomrule
	\end{tabular}
	\vspace{-0.1cm}
\end{table}

\section{Conclusion}
\label{sec:conclusion}
We proposed a graph convolutional network model named WGCN that captures structural features according to nodes' local topologies.
WGCN first obtains nodes' structural fingerprints via a direction and degree aware Random Walk with Restart algorithm,
where the walk is guided by both edge direction and in- and out-degrees of nodes. WGCN then takes the interactions between nodes' structural fingerprints as nodes' structural features.
WGCN also embeds nodes into a latent space to capture nodes' high-order dependencies and latent geometrical relationships. During the message passing, WGCN contextualizes the content and structural features of each node with learnable parameters to navigate the attention-based geometrical aggregation.
Experiments show that WGCN outperforms the baselines by up to 17.07\% in terms of accuracy in transductive node classification on five benchmark datasets.

\section{ACKNOWLEDGMENTS}
\label{sec:ack}
This work is partially supported by Australian Research Council (ARC) Discovery Projects DP180102050. Yunxiang Zhao is supported by the Chinese Scholarship Council (CSC).

\newpage
\bibliographystyle{ACM-Reference-Format}
\balance 
\bibliography{main}
\end{document}